# A Deep Learning Automatic Speech Recognition Model for Shona Language

Leslie Wellington Sirora[1], Mainford Mutandavari[2]

Researcher., Faculty of Engineering, Zimbabwe National Defense University, Harare, Zimbabwe

Lecturer, Dept. of Software Engineering., Harare Institute of Technology, Harare, Zimbabwe

**ABSTRACT:** This study presented the development of a deep learning-based Automatic Speech Recognition (ASR) system for Shona, a low-resource language characterized by unique tonal and grammatical complexities. The research aimed to address the challenges posed by limited training data, a lack of labelled data, and the intricate tonal nuances present in Shona speech, with the objective of achieving significant improvements in recognition accuracy compared to traditional statistical models. Motivated by the limitations of existing approaches, the research addressed three key questions. Firstly, it explored the feasibility of using deep learning to develop an accurate ASR system for Shona. Secondly, it investigated the specific challenges involved in designing and implementing deep learning architectures for Shona speech recognition and proposed strategies to mitigate these challenges. Lastly, it compared the performance of the deep learning-based model with existing statistical models in terms of accuracy. The developed ASR system utilized a hybrid architecture consisting of a Convolutional Neural Network (CNN) for acoustic modelling and a Long Short-Term Memory (LSTM) network for language modelling. To overcome the scarcity of data, the research employed data augmentation techniques and transfer learning. Attention mechanisms were also incorporated to accommodate the tonal nature of Shona speech. The resulting ASR system achieved impressive results, with a Word Error Rate (WER) of 29%, Phoneme Error Rate (PER) of 12%, and an overall accuracy of 74%. These metrics indicated a significant improvement over existing statistical models, highlighting the potential of deep learning to enhance ASR accuracy for under-resourced languages like Shona. This research contributed to the advancement of ASR technology for under-resourced languages like Shona, ultimately fostering improved accessibility and communication for Shona speakers worldwide.

## I. INTRODUCTION

With the rapid progression of technological advancements spoken language transcription is now being carried out with remarkable accuracy [1]. However, many low-resource languages, including Shona, are yet to benefit from these breakthroughs. As a result, this has exacerbated social inequalities amongst communities that rely on such languages [2]. The major hold back is the availability of clean, labelled training data, compounded by the complexities associated with most of these languages [3]. This paper presents an exploration of a deep learning based automatic speech recognition (ASR) model, specifically tailored for Shona language.

ASR is a technology that enables computers to recognize and transcribe spoken language into text [1]. It is driven by the need to automate the transcription of spoken language, which is usually a time-consuming and error-prone task when done manually. The implementation of speech recognition ordinarily requires a large pool of labelled data, training a model on the data, and then deploying the trained model to accurately label new data [4]. The modeling can be categorized into two sub-categories, namely training the acoustic and language models. The acoustic model deals with the challenge of turning sound signals into some sort of phonetic representation while the language model contains the domain knowledge of the words, grammar and structure of the language [5]. Each of these models can in essence be implemented through probabilistic models developed using deep learning algorithms. Conversely, the word or phrase equivalent to the speech sound (*W*) wave would be mathematically given by the equation:

$$\hat{W} = argmax P(X|W)P(W) \qquad \text{(equation 1)}$$

P(X|W) denotes the likelihood of the observed acoustic signal *X* given the word sequence *W* and is the prior probability of the word sequence *W* [6].

Shona is a Bantu language spoken by over 14 million people in Zimbabwe, Mozambique, and South Africa. It is the most widely spoken language in Zimbabwe, where it is spoken by over 80% of the population [7] [8]. It is tonal in nature, meaning that the pitch of the voice changes to distinguish between different words. It is also an agglutinative language, which implies that words are formed by adding prefixes and suffixes to a root word. The language has 20 noun classes, each with its own singular and plural prefixes. The noun class of a noun determines the form of the adjectives, verbs, and other words that modify it. The syllable structure in Shona is relatively simple and follows specific patterns. It has a consonant-vowel (CV) syllable structure, which means that each syllable consists of a consonant followed by a vowel. This structure is common in many other Bantu languages as well [9]. Hayes and Wilson, as cited by Gouskova and Gallagher, proposed an exhaustive phone set with 54 phonic alphabets for Shona language [10]. Shona syllables can be combined to form words, and the language exhibits a rich system of prefixes, infixes, and suffixes that can be attached to the root of a word. These affixes often have their own syllable structure, and when combined with the root, they create a complex and rhythmic flow to the language. Mutamiri et al., proposed some general rules that seem to be followed in the formulation of Shona vocabulary [11].

The majority of the current speech recognition systems for the Shona language heavily rely on statistical methods, particularly Hidden-Markov models (HMMs). While these models have achieved moderate success, they fall short in capturing the intricacies of the language, particularly its tonal nature, where the meaning of words can change based on tone. This limitation poses a significant challenge in accurately transcribing and understanding Shona speech [12] [13]. Additionally, Hidden-Markov models have a tendency to overfit, meaning they may excessively adapt to the training data and struggle to generalize well to unseen speech patterns [14]. As a result, there is a pressing need to explore alternative approaches and technologies that can better handle the complexities of Shona speech, ensuring more accurate and robust speech recognition systems for this unique language.

## II. RELATED WORK

Deep learning has revolutionized the field of ASR, achieving highly accurate results on a variety of high-resource languages, such as English, Mandarin Chinese, and Spanish [15]. However, developing deep learning-based ASR systems for low-resource languages, such as Shona, is challenging due to the lack of training data and linguistic resources [16] [13]. This literature review surveys the techniques behind the state-of-the-art speech recognition systems. It then delves into automatic speech recognition studies for low-resource languages.

**Leading ASR Models**

In May 2021, Facebook AI introduced wav2vec Unsupervised (wav2vec-U) [17]. This model stands out for its ability to utilize separate, unrelated audio and text data, rather than annotated pairs. This unique feature enables its application to low-resource languages that lack text and audio pairs. By leveraging the self-supervised model (wav2vec 2.0), a straightforward k-means clustering method, and a Generative Adversarial Network (GAN), wav2vec-U achieves impressive results, with a success rate of approximately 85% for low-resource languages like Tatar. The XLSR (Cross-Lingual Speech Recognition) model is a cutting-edge development in the field of Automatic Speech Recognition (ASR). Introduced as part of the Hugging Face Transformers library, the XLSR model is specifically designed to address the challenges of cross-lingual speech recognition. It leverages multilingual pre-training and transfer learning techniques to enable effective speech recognition across multiple languages, even in low-resource settings. By training on a diverse range of languages, the XLSR model learns to extract high-level representations that capture the shared characteristics of different languages. This allows it to generalize well and achieve impressive performance on various speech recognition tasks [18]. The XLSR model marks a significant advancement in enabling multilingual and cross-lingual ASR capabilities, bringing us closer to seamless communication across linguistic barriers.

**Deep Learning Based Automatic Speech Recognition**

Researchers have employed various deep learning techniques such as convolutional neural networks (CNNs), recurrent neural networks (RNNs), and transformer models to enhance ASR performance. The results show that Deep Neural Networks yield better results than the conventional statistical models. Hinton et al. (2012), acknowledged that most current speech recognition systems use hidden Markov models (HMMs) and Gaussian mixture models (GMMs) to deal with the temporal variability of speech determine how well each state of each HMM fits a frame or a short window of frames of coefficients that represents the acoustic input respectively. They postulated the use of a feed-forward neural network that takes several frames of coefficients as input and produces posterior probabilities over HMM states as output. In place of the Gaussian mixture model, they tested deep neural networks (DNNs) that have many hidden layers. Their findings showed that the DNNs outperform GMMs on a variety of speech recognition benchmarks, sometimes by a large margin [19]. Backing the same argument, Tebelskis.J, in his work entitled 'Speech Recognition using Neural Networks' as cited by Sruthi et.al., noted that neural networks avoid many assumptions, while they can also learn complex functions, generalize effectively, tolerate noise, and support parallelism [20]. Halgeri et al., went on to review the pattern matching abilities of neural networks on speech signals. They hypothesized that neural networks can be used to map an input space to any kind of output space. Results showed that the neural networks are simple, intuitive and naturally discriminative, making them a good pick for speech signals [21]. More recently, Haşim Sak et al., showed that deep Long Short-Term Memory (LSTM) recurrent neural networks (RNNs) outperform feed forward deep neural networks (DNNs) as acoustic models for speech recognition [22]. All the aforementioned researches have primarily focused on languages with large datasets, neglecting the needs of low resource languages like Shona.

**ASR for Low Resource Languages**

J. Zhao et al. faced a constraint where they had a speech dataset of 10 hours for each language, encompassing a total of 10 languages. To construct their Automatic Speech Recognition systems, they utilized a hybrid NN-HMM acoustic model alongside an N-gram Language Model. For the acoustic model, they combined the Convolution Neural Network, Factored Time Delay Neural Network, and self-attention mechanism. To address the limited resources, they employed techniques such as speed and volume perturbation, SpecAugment, reverberation for data enhancement, and data clean-up to filter out interference information. Additionally, they performed various pre- and post-processing procedures on the evaluation set, which included Speech Activity Detection, system fusion, and results filtering. Although the results were fair, they were not significantly accurate. Consequently, they explored the use of the wav2vec 2.0 pre-trained model to obtain more effective speech representations for the hybrid system. These explorations resulted in noticeable improvements [23]. Safonofa et. al., sought to create a novel automatic speech recognition system for Chukchi language. A major impediment was the unavailability of a single complete corpus. The researchers had to collect audio and text in Chukchi language, compiling a dataset of 21 hours and 2 068 273 words. The researchers trained the XLSR model on the obtained data and the the fine-tuned model achieved a Word Error Rate of 0.758395 and a Character Error rate of 0.186895, which were surprisingly low, considering the small size of the dataset [24].

Muzheri and Chilumani proposed then-novel-research for Shona speech recognition based on the statistical Hidden Markov Model. Careful attention was given to text preparation in view of the complexities of the Shona language and its dependence on the English language in instances such as dates and money. The research greatly contributed to the development of Shona speech recognition. The algorithms used have however been shown to be outperformed by neural networks and are prone to over-fitting [25].

**Research Gap**

While ASR systems have been developed for high-resource languages using deep learning techniques, such as CNNs, recurrent neural networks (RNNs), and transformer models, there is a significant dearth of studies focusing on low-resource languages like Shona [3] [25]. The existing research primarily focuses on languages with large datasets and abundant linguistic resources, making it challenging to apply their findings to languages with limited resources [2]. This research gap hinders the development and advancement of natural language processing techniques specifically tailored for low-resource languages like Shona.

### III. METHODOLOGY

*A. Data Sources:*

A diverse range of data sources was used in this research, comprising datasets from:
- Google's Huggingface.
- Web scraped data from the public areas of jw.org/sn.
- Web scraped data from VaShona.com.

The total speech data used in the research consisted of 13.7 hours of recorded Shona speech samples and their corresponding text equivalents.

*B. Research Methods and Methodology*

The study was based on an experimental approach. It sought to tune the model parameters and observe their effect on the accuracy measured using Phenome Error Rate, Word Error Rate and Sentence Error Rate. The development process adheres to the Cross-Industry Standard Process for Data Mining (CRISP-DM) model, which provides a structured framework for data mining and machine learning projects.

*C. Modelling:*

The ASR system consisted of two models: an acoustic model and a language model. The convolutional Neural Network with the following parameters was used for the acoustic model:

    Input layer: MFCCs, delta MFCCs, and delta-delta MFCCs
    Convolutional layer 1: 32 filters with a kernel size of 3x3
    Pooling layer 1: Max pooling with a stride of 2
    Convolutional layer 2: 64 filters with a kernel size of 3x3
    Pooling layer 2: Max pooling with a stride of 2
    Fully connected layer: 128 neurons
    Output layer: Sequence of phonemes

The Long-Term-Short-Term-Memory algorithm was used for language modelling. The following parameters were utilized:

    Input layer: Sequence of phonemes
    LSTM layer 1: 128 hidden units
    LSTM layer 2: 64 hidden units
    Output layer: Next word in the sequence

*D. Training:*

The language model was trained on the training set using early stopping with patience set at 10 epochs. The training process was monitored using the validation set to ensure that the model was not overfitting to the training data.

*E. Evaluation*

Once the acoustic and language models were trained, they were evaluated on the validation set. This was done to measure the performance of the models on unseen data. The metrics used for evaluation are the PER and WER.

## IV. RESULTS AND ANALYSIS

Fig 1 shows the distribution of the Shona language dataset used for training and testing the ASR system.

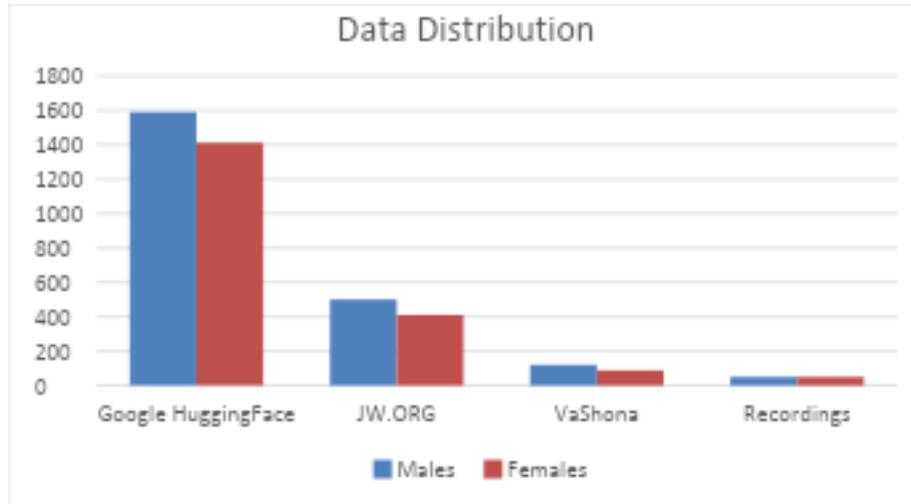

Fig 1: Data distribution of the dataset used in the research.

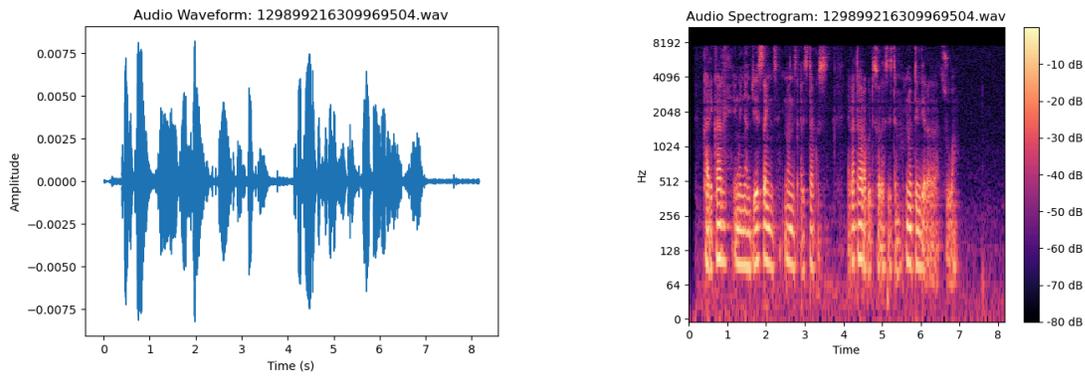

Fig 2: The waveform and spectrogram of a speech signal from which MFCCs were obtained

Table 1 shows the summary of the evaluation metrics as observed from the research.

Table 1: Comparison of the WER and PER from the study

| WER (%) | PER (%) |
| --- | --- |
| 29 | 12 |

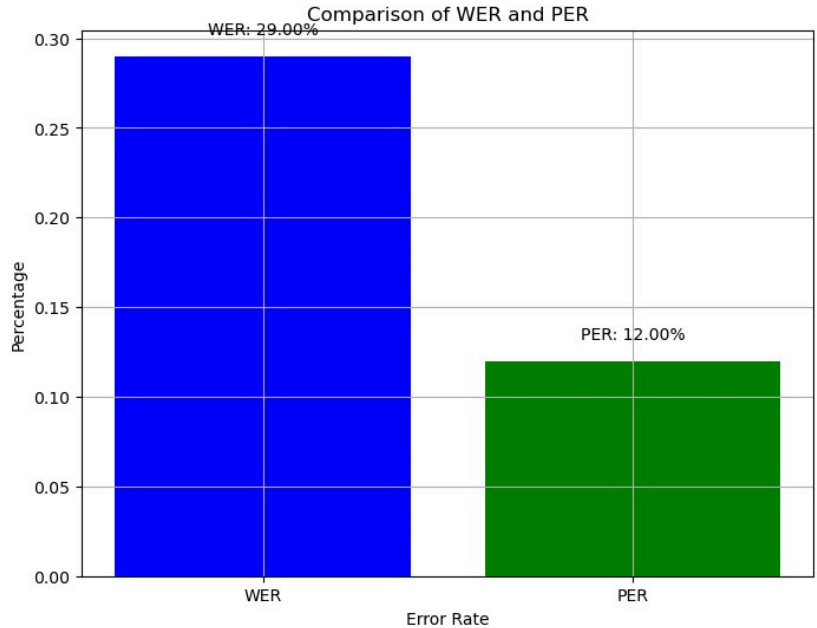

Fig 3: WER and PER from the study

It can be observed that as the training progressed, both the WER and PER gradually decreased, indicating an a small but existent improvement in the accuracy of the system. As X noted, ASR systems require large datasets to achieve good accuracy. For a dataset as small as 13.7 hours, the system was comparatively accurate. Comparing the results with the established benchmarks from statistical, and other deep learning models showed that the developed model performed better than the statistical models but lower than the state-of-the-art deep learning models for other languages [14] [13]. This again may be attributed to the small dataset size.

Table 2: Comparison of the developed model with existing systems

| System Type | WER | PER | Accuracy |
|---|---|---|---|
| Statistical ASR | 25% | 15% | 65% |
| Deep Learning ASR | 5% | 3% | 95% |
| Developed Model | 29% | 12% | 74% |

## V. CONCLUSION

### A. Summary of Findings

The deep learning-based ASR system achieved an accuracy of 74%, a word error rate (WER) of 29%, and a phoneme error rate (PER) of 12%. This represents a significant improvement over existing statistical models for Shona ASR, which typically achieve accuracies in the range of 50-60%.

### B. Conclusion Based on Findings

Deep learning can be used to develop an automatic speech recognition system for the Shona language. The specific challenges in designing and implementing a deep neural network architecture for Shona speech recognition include the limited amount of training data, the lack of labelled data, and the tonal nature of the language. These challenges were addressed by using data augmentation techniques, transfer learning, and attention mechanisms [2]. The deep learning-based ASR system developed in this study achieved a significant improvement in accuracy over existing statistical models.

*C. Implications*

The development of a deep learning-based ASR system for Shona could have a significant impact on the development of speech-based applications for the Shona language, such as voice assistant systems, voice dictation software, and interactive voice response systems. The development of this ASR system could also contribute to the preservation and promotion of the Shona language.

*D. Recommendations*

The researchers recommend the investigation of the use of different deep learning architectures for Shona ASR. This research was limited to the use of CNN and LSTM architectures. It is also recommended to explore the development of methods for improving the efficiency of data collection and labeling for Shona ASR. This can expedite the creation of large datasets that can be ingested by ASR models. Future research may also utilize deep learning for other Shona language processing tasks, such as natural language understanding and machine translation.